\documentclass{article}
\usepackage{spconf,amsmath,hyperref}
\usepackage{amsmath}
\usepackage{graphicx}
\usepackage{amssymb}
\usepackage{booktabs}
\usepackage{array}
\usepackage{pifont}
\newcommand{\cmark}{\ding{51}}  
\newcommand{\xmark}{\ding{55}}  
\usepackage{adjustbox}  
\usepackage[table]{xcolor}
\usepackage{cleveref}
\usepackage{makecell} 
\usepackage{tabularx}
\usepackage{array,makecell,booktabs,xcolor}
\usepackage{subcaption} 
\usepackage[skip=6pt]{caption}
\usepackage{booktabs,array}

\title{HAVT-IVD: Heterogeneity-aware Cross-modal Network for Audio-Visual Surveillance: Idling Vehicles Detection with Multichannel Audio and Multiscale Visual Cues}
\name{Author(s) Name(s)\thanks{Thanks to XYZ agency for funding.}}
\address{Author Affiliation(s)}
\name{Xiwen Li$^{\star}$ \qquad Xiaoya Tang$^{\star}$ \qquad Tolga Tasdizen$^{\star\dagger}$}
  
\address{$^{\star}$ Scientific Computing and Imaging Institute, Salt Lake City, USA \\
  $^{\dagger}$ Department of Electrical and Computer Engineering, University of Utah, Salt Lake City, USA}
\begin{document}
%
\maketitle
\begin{abstract}
Idling vehicle detection (IVD) uses surveillance video and multichannel audio to localize and classify vehicles in the last frame as \emph{moving}, \emph{idling}, or \emph{engine-off} in pick-up zones. IVD faces three challenges: \textbf{(i)} modality heterogeneity between visual cues and audio patterns; \textbf{(ii)} large box scale variation requiring multi-resolution detection; and \textbf{(iii)} training instability due to coupled detection heads. The previous end-to-end (E2E) model~\cite{Li2024JointAI} with simple CBAM-based \cite{woo2018cbam} bi-modal attention fails to handle these issues and often misses vehicles.  
We propose \textbf{HAVT-IVD}, a heterogeneity-aware network with a visual feature pyramid and decoupled heads.  
Experiments show HAVT-IVD improves mAP by \textbf{7.66} over the disjoint baseline and \textbf{9.42} over the E2E baseline.

\end{abstract}

\section{Introduction}
\label{sec:intro}
\begin{figure*}[htbp]
    \centering
    \includegraphics[width=\linewidth]{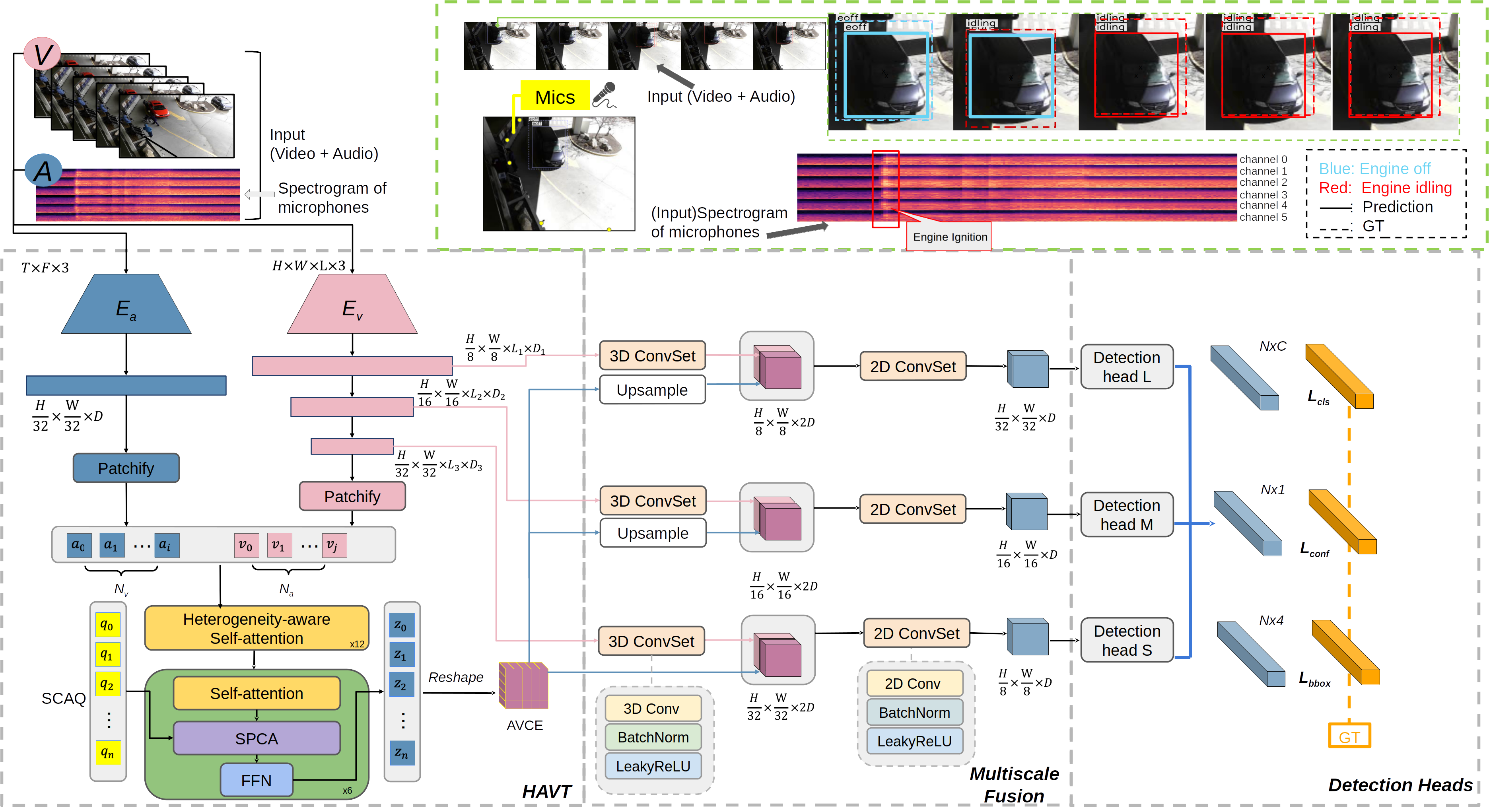}
    \caption{HAVT-IVD architecture. Shapes not to scale.}
    \label{fig:algorithm-pipeline}
\end{figure*}

Idling vehicles with running engines emit pollutants, waste fuel, and cause engine wear, making IVD algorithms essential for detecting and monitoring them. Thermal-based methods \cite{Bastan2018RemoteDO} infer idling status from vehicle heat-up/cool-down patterns but suffer from high latency due to slow thermal image acquisition. Instead, leveraging portable microphones and web cameras within a surveillance setup offers a practical and scalable solution. Microphone arrays are a natural choice for portable audio-visual IVD. However, compact synchronized MEMS arrays combined with classical sound-source localization (e.g., GCC-PHAT, TDOA, FOA) often fail to deliver sufficient array/beamforming gain at long stand-off distances, where idling engines are low-amplitude and spectrally narrow compared to high-frequency acceleration events \cite{MM-Distill}. \cite{Li2023RealTimeIV} therefore prefer a distributed layout of \textbf{unsynchronized} wireless microphones placed closer to typical stopping zones, which raises per-channel proximity SNR without requiring tight array synchronization—making it more suitable for IVD. It enables IVD to be formulated as an audio-visual vehicle status detection problem. To be specific, using synchronized video from a surveillance camera and spectrograms from six evenly spatially spaced microphones (\cref{fig:algorithm-pipeline} green box), IVD localizes each vehicle in the last frame and classifies its state as \emph{moving}, \emph{idling}, or \emph{engine-off}. 

Designed for IVD, Real-Time IVD \cite{Li2023RealTimeIV} uses a disjoint pipeline requiring daily user intervention, causing errors and poor scalability; AVIVDNet \cite{Li2024JointAI} moves to an E2E audio-visual detector with CBAM-inspired attention \cite{woo2018cbam}, but its deconvolutional upsampling to the visual resolution plus simple attention limits cross-modal routing and fails to capture heterogeneity. More broadly, audio-visual learning tasks such as sounding object segmentation \cite{hao2024improving, zhou2022avs, gao2023avsegformer} and active speaker detection \cite{wang2023loconet}; most use mono audio and thus cannot exploit multi-channel spatial cues or jointly infer detection and state. Audio-visual knowledge distillation \cite{LiuSMDAFAS, gan2019self, riverahurtado2021mmdistillnet, vasudevan2020semantic} transfers visual supervision to audio (e.g., \cite{gan2019self} detects moving vehicles with binaural microphones) but emphasizes modality transfer rather than joint representation learning. We propose HAVT-IVD, an E2E model that learns from multi-channel audio and video jointly, specialized for IVD in complex scenes.

\begin{figure}[htbp]
    \centering
    \includegraphics[width=\linewidth]{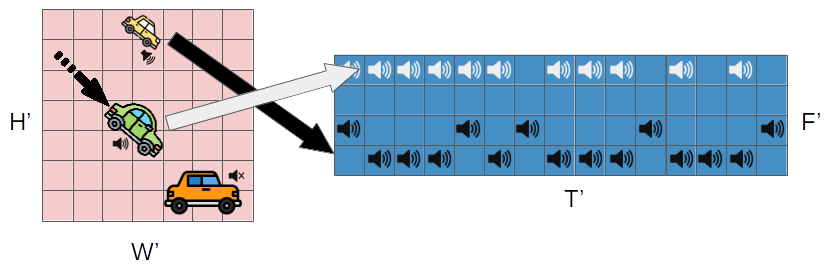}
    \caption{Illustration of Instance Heterogeneity}
    \label{fig:heterogeneity}
\end{figure}

Three untackled key challenges remain for IVD:  
\textbf{(1)} \textbf{Modality-heterogeneity cues.} As shown in \cref{fig:heterogeneity}, due to a cross-modal spatial distribution mismatch, the model must select the correct audio evidence (high/low-rpm from high/low-frequency bands) for each individual vehicle feature. \textbf{(2)} \textbf{Scale variation.} Objects near the camera appear large while distant vehicles are only a few pixels, so effective IVD needs multi-resolution features and scale-aware heads. \textbf{(3)} \textbf{Coupled detection head.} Sharing weights across different objectives leads to gradient conflicts. To address these challenges, we propose \textbf{HAVT-IVD}, a network that performs global audio–visual alignment via flexible self-attention to mitigate heterogeneity, fuses multi-scale visual features to improve detection of vehicles at varying distances, and employs decoupled heads to separate classification from localization to mitigate gradient conflict. Our contributions are twofolds: 
\begin{itemize}
    \setlength{\itemsep}{0pt} 
    \setlength{\parskip}{0pt} 
  \setlength{\parsep}{0pt}  
    \item We propose \textbf{HAVT-IVD}, a heterogeneity-aware network that incorporates feature pyramids and decoupled heads to better address IVD feature heterogeneity.
    \item Extensive experiments on the AVIVD dataset \cite{Li2024JointAI} and MAVD \cite{MM-Distill} validate the effectiveness of HAVT-IVD on IVD and general audio–visual vehicle detection tasks.
\end{itemize}

\section{Proposed Method}
\subsection{Problem Formulation and AVIVD Dataset}
AVIVD \cite{Li2024JointAI} was collected with a remote surveillance camera and evenly spaced wireless microphones deployed roadside at a hospital pick-up zone. Both modalities continuously and non-intrusively monitor the target area. Each video–audio pair is annotated with bounding boxes and class labels (\emph{moving}, \emph{idling}, \emph{engine-off}) for every visible vehicle in the drive lanes.  
We define IVD as follows: given a video clip
\(
V \in \mathbb{R}^{D \times C \times H \times W}
\)
and audio
\(
A \in \mathbb{R}^{M \times T \times F},
\)
the model predicts bounding boxes \((\text{bbox})\) and class labels \((\text{cls})\) for each vehicle in \(V\).  
Here, \(D\) is the number of frames with \(C\) channels of spatial size \(H \times W\);  
\(M\) the number of microphones; and \(T,F\) the time and frequency axes in the STFT domain.  
Applying an STFT to each of the \(M = 6\) audio channels yields a 2-D spectrogram \(A\), where each pixel encodes the signal power at a given time (horizontal axis) and frequency (vertical axis), as shown in \cref{fig:algorithm-pipeline}.

\subsection{Heterogeneity-aware Audio-Visual Transformer}
\label{sec:transformer}
AVIVDNet \cite{Li2024JointAI} aligns audio features to the spatial resolution of visual features via a deconvolution layer and fuses them using an attention module inspired by CBAM \cite{woo2018cbam}.  
Such a forced spatial alignment (i) distorts audio cues and (ii) limits the model’s ability to flexibly route patches across modalities.  
Instead, we process raw audio and visual feature patches directly, preserving the integrity of the audio representation.  

In IVD, we observe a pronounced \emph{audio--visual feature evidence heterogeneity}: as illustrated in \cref{fig:heterogeneity}, video encodes \emph{where} vehicles are, while audio encodes \emph{when} and \emph{how strongly} engines are active, making the two modalities inherently complementary but misaligned.  
For each vehicle in the last frame, the model must identify the most relevant audio evidence from the entire clip, since different vehicles may correspond to distinct portions of the audio mixture.  
This selection is class-specific: idling relies on low-frequency narrow-band rumbles, moving on broadband transients temporally aligned with motion, and engine-off on the near absence of engine sound. As shown in \cref{fig:algorithm-pipeline}, \textbf{HAVT-IVD} applies a 3D CNN visual encoder $E_v$ and an audio encoder $E_a$ to the video $V \!\in\! \mathbb{R}^{D \times C \times H \times W}$ and audio $A \!\in\! \mathbb{R}^{M \times T \times F}$, yielding downsampled feature maps $ \mathbb{R}^{D \times \frac{H}{32} \times \frac{W}{32}} $ and $ \mathbb{R}^{D \times \frac{T}{32} \times \frac{F}{32}} $ that are then \emph{patchified} into $N_v$ video patches (red in \cref{fig:algorithm-pipeline}, indexed by $j$) and $N_a$ audio patches (blue, indexed by $i$). Since interchannel audio cues drive localization, each time frequency patch \(a_i\) encodes cross channel energy, enabling engine localization for visually static vehicles.

This design first performs global alignment and then spatial aggregation, jointly addressing positional mismatch and semantic drift, yielding a fused representation \emph{AVCE} (purple in \cref{fig:algorithm-pipeline}) that supports stable multi-scale fusion and decoupled heads.  
To mitigate heterogeneity, the \(N_v + N_a\) patches are concatenated and jointly fed into a 12-layer self-attention encoder ($\operatorname{fsa}$: self-attention layers), $\operatorname{fsa}\!\bigl(\{v_0,\dotsc,v_j, a_0,\dotsc,a_i\}\bigr)$
to learn global routing. This provides flexible, content-adaptive alignment: visual motion and audio semantic patches exchange context in token space, reducing cross-modal semantic and distribution mismatch while preserving the input–output shape and modality-aware information.  
Afterward, Spatial-Pulling Cross-Attention (SPCA) layers use \emph{SCAQ (Spatial Cross-Modal Aggregation Queries)}, grid-aligned query slots that select and pool evidence from the shared audio–visual key–value memory and reshape the aggregated tokens into a \(7{\times}7{\times}D\) spatial \emph{AVCE} feature map aligned with the visual grid.


\subsection{Feature Pyramid and Decoupled Heads}
To overcome scale variation, we use AVCE to fuse features across the visual pyramid, matching each box to the most appropriate scale. As shown in \cref{fig:algorithm-pipeline}, the 3D visual encoder \(E_v\) outputs feature maps at large \((\frac{H}{8}\!\times\!\frac{W}{8}\!\times\!L_{1}\!\times\!D_{1})\), medium \((\frac{H}{16}\!\times\!\frac{W}{16}\!\times\!L_{2}\!\times\!D_{2})\), and small \((\frac{H}{32}\!\times\!\frac{W}{32}\!\times\!L_{3}\!\times\!D_{3})\) spatial resolutions. We apply \(1\!\times\!1\!\times\!L\) 3D convolutions to remove the temporal dimension and unify channels to \(D\). The AVCE output is then upsampled to match each resolution and concatenated with the corresponding visual feature map along channels, yielding three fused tensors of shape \(\frac{H}{8}\!\times\!\frac{W}{8}\!\times\!2D\), \(\frac{H}{16}\!\times\!\frac{W}{16}\!\times\!2D\), and \(\frac{H}{32}\!\times\!\frac{W}{32}\!\times\!2D\). A \(1\!\times\!1\) 2D convolution reduces each fused map back to \(D\) channels for multi-scale detection.

To mitigate gradient conflict between joint localisation and state classification, we adopt a YOLO-style \emph{decoupled head}: each fused pyramid level feeds an independent detection head with two \(3{\times}3\) convolutional branches—one for \(C\) class logits, the other for box parameters and objectness. Each cell thus outputs \(C+1+4\) values, and each level predicts \(N \!\in\! \{28^{2}, 14^{2}, 7^{2}\}\) bounding boxes. Training jointly supervises objectness confidence, vehicle-state classification, and bounding-box regression following YOLOv5: $L_{\text{total}} = \lambda_{conf}L_{\text{conf}} + \lambda_{cls}L_{\text{cls}} + \lambda_{reg}L_{\text{bbox}} \label{eq:total_loss}$.
We set $\lambda_{conf} = 1$, $\lambda_{cls} = 1$, and $\lambda_{reg} = 5$ to balance the three terms.
\section{Experiments}
\textbf{Datasets.} We evaluate on AVIVD\cite{Li2024JointAI} and MAVD\cite{MM-Distill}. AVIVD, built for audio and visual IVD, has 76,490 training and 8,431 disjoint test pairs; boxes (M/I/Eoff) are 26,924/36,968/41,868 in training and 2,908/2,669/3,422 in testing. MAVD is collected with an in vehicle multichannel acoustic camera for street vehicle detection.\\
\textbf{Metrics.} We evaluate model performance using mean average precision (mAP) and average precision (AP), the standard metrics for action detection, with IoU threshold set to $0.5$.\\
\textbf{Implementation Details.} Inputs are $224\times224$ with 16 frame clips. Audio uses a 5 s six channel segment centered on the last frame at $48{,}000$ Hz, converted to mel spectrograms (window 1024, hop 512, 128 bins) yielding $128\times469$ per channel; We train PyTorch models on NVIDIA A6000 with a batch size of 16, a learning rate of $1\times10^{-3}$, for up to 100 epochs with early stopping (patience 50).
\begin{table}[!t]
  \centering
  \small
  \setlength{\tabcolsep}{3.5pt}
  \renewcommand{\arraystretch}{1}
\caption{Comparisons on AVIVD and adapted AVSBench models (bold marks best, underline second best).}
  \label{tab:overall_compact}
  \resizebox{\columnwidth}{!}{%
  \begin{tabular}{l c l c c c c}
    \toprule
    Method & E2E & \makecell{Audio \\ Backbone} & mAP & AP(M) & AP(I) & AP(Eoff)\\
    \midrule
    \multicolumn{7}{l}{\textit{(A) AVIVD methods}}\\
    Real-Time IVD \cite{Li2023RealTimeIV} & \xmark & R50 (frozen) & 80.97 & 92.45 & 68.93 & 81.55\\
    Feature Concat. & \cmark & MNv3 & 77.45 & \underline{93.97} & 60.35 & 78.02\\
    Feature Concat. & \cmark & R50 (frozen) & 77.35 & 93.67 & 66.19 & 72.18\\
    AVIVDNet & \cmark & MNv3 & 78.89 & 90.77 & 66.81 & 79.10\\
    AVIVDNet\cite{Li2024JointAI} & \cmark & R50 (frozen) & 79.21 & 93.43 & 66.74 & 77.47\\
    \hline
    HAVT-only & \cmark & MNv3 & 80.95 & 85.25 & 73.19 & 84.41\\
    HAVT-IVD-SC (Ours) & \cmark & MNv3 & \underline{85.28} & 88.14 & \underline{80.19} & \underline{87.51}\\
    \textbf{HAVT-IVD (Ours)} & \cmark & MNv3 & \textbf{88.63} & \textbf{94.35} & \textbf{83.41} & \textbf{88.12}\\
    \hline
    \hline
    \addlinespace[2pt]
    \multicolumn{7}{l}{\textit{(B) AVSBench models adapted to AVIVD (encoders unified to MobileNetV2)}}\\
    TPAVI (ECCV'22) \cite{zhou2022avs} & \cmark & MNv3 & 23.27 & 38.66 & 4.21 & 26.94\\
    AVSegFormer (AAAI'23) \cite{gao2023avsegformer} & \cmark & MNv3 & 14.65 & 31.12 & 0.07 & 12.77\\
    ECMVAE (ICCV'23) \cite{Mao2023MultimodalVA} & \cmark & MNv3 & 13.38 & 38.60 & 0.72 & 0.83\\
    AVSBiGen (AAAI'24) \cite{hao2024improving} & \cmark & MNv3 & 23.54 & 31.40 & 3.20 & 36.03\\
    \textbf{HAVT-IVD (Ours)} & \cmark & MNv3 & \textbf{88.63} & \textbf{93.45} & \textbf{83.41} & \textbf{88.12}\\
    \bottomrule
  \end{tabular}
  }
\end{table}

\subsection{Quantitative Comparison with Prior Arts}
\noindent\textbf{Comparison with IVD methods.}
We evaluate heterogeneity mitigation on AVIVD. As shown in \cref{tab:overall_compact} \textit{(A)}, \textbf{HAVT-IVD} attains the best result with \textbf{88.63,mAP@0.5}. Compared with the three stage \emph{Real-Time IVD} pipeline (already aided by heuristic mic selection), it gains \textbf{+7.66 mAP} and lifts the \emph{Idling} category by \textbf{+14.5 AP} (83.41 vs.\ 68.93). The two \emph{Feature Concat.} baselines show that naive stacking of audio and visual features hurts performance, with mAP dropping to 77 and \emph{Idling} AP below 61. Replacing concatenation with AVIVDNet’s bidirectional attention recovers 1.5 mAP but still trails the transformer on \emph{Idling} (83.41 vs.\ 66.81). Three \emph{HAVT} variants also outperform AVIVDNet, especially on AP(I) and AP(Eoff), highlighting the benefit of flexible patch routing via self attention, which reasons about cross modal heterogeneity better than audio feature conversion and simple CBAM-style bidirectional attention.\\
\noindent\textbf{Comparison with SOTAs from related tasks (AVSBench).} We compare HAVT-IVD with state of the art audio-visual video segmentation (AVS) models, the closest available methods for our setting because they take both audio and video to locate and segment sounding objects. Methods without publicly available code are excluded. For fairness and to avoid overfitting, we replace their encoders with the same lightweight MobileNetV3 used in HAVT-IVD, append convolutional layers to downsample outputs from $224 \times 224$ to $7 \times 7$, and attach a single detection head to adapt them to our task. As shown in \cref{tab:overall_compact} \textit{(B)}, HAVT-IVD achieves a large gain, clearly surpassing the best AVSBench baseline (TPAVI, $23.27$ mAP). This significant margin highlights the advantage of our heterogeneity aware design for this specialized cross modal detection problem.

\begin{table}[htbp]
  \centering
  \scriptsize
  \setlength{\tabcolsep}{6pt}
  \renewcommand{\arraystretch}{0.95}
  \caption{Ablations on AVIVD. Light gray: best setting. (bold marks best)}
  \label{tab:ablation_unified}
  \begin{tabular}{l l c c c c}
    \toprule
    Study & Setting & mAP & AP(M) & AP(I) & AP(Eoff)\\
    \midrule
    \multicolumn{6}{l}{\textit{(A) Multiscale Visual Fusion}}\\
    & $7{\times}7$                         & 85.28 & 88.14 & 80.19 & 87.51 \\
    & $7{\times}7,\,14{\times}14$          & 82.40 & 88.10 & 67.29 & \textbf{91.81} \\
    \rowcolor{gray!30}
    & $7{\times}7,\,14{\times}14,\,28{\times}28$ & \textbf{88.63} & \textbf{94.35} & \textbf{83.41} & 88.12 \\
    \hline
    \addlinespace[2pt]
    \multicolumn{6}{l}{\textit{(B) SCAQ/AVCE Resolution}}\\
    \rowcolor{gray!30}
    & $N_{SCAQ}{=}49$ ($7{\times}7$)   & \textbf{88.63} & 94.35 & \textbf{83.41} & \textbf{88.12} \\
    & $N_{SCAQ}{=}196$ ($14{\times}14$)& 85.06 & \textbf{96.37} & 73.61 & 85.19 \\
    & $N_{SCAQ}{=}784$ ($28{\times}28$)& 85.25 & 96.00 & 74.74 & 85.01 \\
    \hline
    \addlinespace[2pt]
    \multicolumn{6}{l}{\textit{(C) Detection Head}}\\
    & Coupled    & 80.95 & 85.25 & 73.19 & 84.41 \\
    \rowcolor{gray!30}
    & Decoupled  & \textbf{85.28} & \textbf{88.14} & \textbf{80.19} & \textbf{87.51} \\
    \hline
    \addlinespace[2pt]
    \multicolumn{6}{l}{\textit{(D) Number of Microphones}}\\
    & 1 & 67.98 & 82.96 & 51.78 & 69.20 \\
    & 3 & \textbf{80.98} & \textbf{87.08} & 72.96 & 82.91 \\
    & 6 & 80.95 & 85.25 & \textbf{73.19} & \textbf{84.41} \\
    \bottomrule
  \end{tabular}
\end{table}

\subsection{Ablation Studies}
We ablate multiscale feature fusion, AVCE resolution, decoupled heads, and microphone number.\\
\textbf{Effectiveness of Multiscale Fusion. \cref{tab:ablation_unified} \textit{(A)}.} Incorporating multiscale visual features significantly improves detection performance compared to using a single-scale feature map. While the two-scale setting ($7\times7$, $14\times14$) improves AP(Eoff) to $91.81$, it leads to a drop in AP(I) due to limited semantic richness. By further adding the $28\times28$ resolution, the full multiscale setup achieves the best overall performance with an mAP of $88.63$, outperforming the single-scale baseline by $3.35$. These results demonstrate the importance of combining both low-resolution semantic and high-resolution spatial cues for accurate box detection.
\newline
\textbf{SCAQ/AVCE Spatial Resolution. \cref{tab:ablation_unified} \textit{(B)}.} We examine the number of SCAQ set that drives SPCA to form the AVCE (\cref{sec:transformer}). We evaluate \(N_{SCAQ}=49,196,784\) corresponding to \(7{\times}7,14{\times}14,28{\times}28\) grids, with AVCE features up- or downsampled for downstream compatibility. Increasing \(N_{SCAQ}\) enhances the model’s ability to pull fine-grained evidence into the spatial domain, but also raises sensitivity to irrelevant local features and risk of overfitting. The \(7{\times}7\) setting offers the best trade-off, achieving 88.63 mAP overall and excelling on AP(I) and AP(Eoff) for idling and engine-off vehicles. These results suggest that a compact SCAQ set provides sufficient spatial-pulling expressiveness while maintaining robust generalization for IVD.
\newline
\textbf{Coupled V.S. Decoupled Detection Head. \cref{tab:ablation_unified} \textit{(C)}.} Decoupled head surpasses the Coupled head on all metrics, boosting mAP from $80.95$ to $85.28$ and improving idle detection by $+7.00$ AP.  
This confirms that decoupling enhances feature separation and yields better detection performance.
\newline
\textbf{Robustness to Microphone Numbers. \cref{tab:ablation_unified} \textit{(D)}.} Ablating microphone count with 1, 3, and 6 channels shows that one microphone performs worst, with AP(I) more than 21 points lower than with multiple microphones. Reducing from 6 to 3 causes only minor drops, 0.23 on AP(I) and 1.5 on AP(Eoff), demonstrating the robustness of HAVT to microphone numbers and making three microphones a practical choice when space is constrained.

\subsection{Model Applicability on MAVD}
Our heterogeneity-aware model also works well on MAVD \cite{MM-Distill} (\cref{tab:mavd}), whose in-vehicle multi-channel acoustic–camera setup closely resembles a self-driving scenario. Benefiting from encoder self-attention and SCAQ-driven cross-attention, HAVT-IVD boosts performance over AVIVDNet, reaching 69.86 mAP@Avg and 84.03 mAP@0.5, and still achieving 55.69 mAP at IoU 0.75, demonstrating strong generalization to MAVD and practical applicability to real-world vehicle detection.
\begin{table}[htbp]
    \centering
    {\small  
    \renewcommand{\arraystretch}{1.1}
    \setlength{\tabcolsep}{3pt}  

    \caption{Performance on MAVD Day Data.}
    \label{tab:mavd}
 \resizebox{\linewidth}{!}{  
    \begin{tabular}{
        |c|c|c|c|c|
    }
    \hline
    Method & KD & $mAP@Avg$ & $mAP@{0.5}$ & $mAP@{0.75}$ \\
    \hline
    StereoSoundNet \cite{Gan2019SelfSupervisedMV}  & \cmark & $44.05$ & $62.38$ & $41.46$ \\
    Pairwise Loss \cite{Liu2019StructuredKD}  & \cmark & $40.45$ & $59.72$ & $36.73$\\
    AFD Loss \cite{Wang2020PayAT}  & \cmark & $44.27$ & $62.00$ & $41.90$ \\
    MM-DistillNet \cite{MM-Distill}  & \cmark & $44.58$ & $62.66$ & $42.39$\\
    AVD Loss \cite{TowardsRobustAudioBasedVehicleDetectio}  & \cmark & $\underline{58.39}$ & $\underline{78.91}$ & $\textbf{56.29}$\\
    \hline
    AVIVDNet \cite{Li2024JointAI} & \xmark & $35.75$ & $55.41$ & $16.08$ \\
    HAVT-IVD (Ours) & \xmark & $\mathbf{69.86}$ & $\textbf{84.03}$ & $\underline{55.69}$ \\
    \hline
    \end{tabular}
    }}
\end{table}

\section{Conclusion}
In this paper, we present HAVT-IVD, a heterogeneity-aware transformer-based model that integrates a feature pyramid and decoupled prediction heads for IVD. Extensive experiments on the AVIVD and MAVD datasets show that HAVT-IVD is a strong and extensible cross-modal feature learner. For future work, we plan to reformulate IVD as a pure classification problem, eliminating the detection step, in line with prevailing approaches for surveillance-style data.

\vfill\pagebreak




\bibliographystyle{IEEEbib}
\bibliography{strings,refs}

\begin{thebibliography}{10}

\bibitem{Li2024JointAI}
Xiwen Li, Rehman Mohammed, Tristalee Mangin, Surojit Saha, Ross~T. Whitaker, Kerry Kelly, and Tolga Tasdizen,
\newblock ``Joint audio-visual idling vehicle detection with streamlined input dependencies,''
\newblock {\em ArXiv}, vol. abs/2410.21170, 2024.

\bibitem{woo2018cbam}
Sanghyun Woo, Jongchan Park, Joon-Young Lee, and In~So Kweon,
\newblock ``Cbam: Convolutional block attention module,''
\newblock in {\em Proceedings of the European conference on computer vision (ECCV)}, 2018, pp. 3--19.

\bibitem{Bastan2018RemoteDO}
Muhammet Bastan, Kim-Hui Yap, and Lap-Pui Chau,
\newblock ``Remote detection of idling cars using infrared imaging and deep networks,''
\newblock {\em Neural Computing and Applications}, vol. 32, pp. 3047 -- 3057, 2018.

\bibitem{MM-Distill}
Francisco~Rivera Valverde, Juana~Valeria Hurtado, and Abhinav Valada,
\newblock ``There is more than meets the eye: Self-supervised multi-object detection and tracking with sound by distilling multimodal knowledge,''
\newblock {\em 2021 IEEE/CVF Conference on Computer Vision and Pattern Recognition (CVPR)}, pp. 11607--11616, 2021.

\bibitem{Li2023RealTimeIV}
Xiwen Li, Tristalee Mangin, Surojit Saha, Rehman Mohammed, Evan Blanchard, Dillon Tang, Henry Poppe, Ouk Choi, Kerry Kelly, and Ross Whitaker,
\newblock ``Real-time idling vehicles detection using combined audio-visual deep learning,''
\newblock in {\em Emerging Cutting-Edge Developments in Intelligent Traffic and Transportation Systems}, pp. 142--158. IOS Press, 2024.

\bibitem{hao2024improving}
Dawei Hao, Yuxin Mao, Bowen He, Xiaodong Han, Yuchao Dai, and Yiran Zhong,
\newblock ``Improving audio-visual segmentation with bidirectional generation,''
\newblock in {\em Proceedings of the AAAI conference on artificial intelligence}, 2024, vol.~38, pp. 2067--2075.

\bibitem{zhou2022avs}
Jinxing Zhou, Jianyuan Wang, Jiayi Zhang, Weixuan Sun, Jing Zhang, Stan Birchfield, Dan Guo, Lingpeng Kong, Meng Wang, and Yiran Zhong,
\newblock ``Audio-visual segmentation,''
\newblock in {\em European Conference on Computer Vision}, 2022.

\bibitem{gao2023avsegformer}
Shengyi Gao, Zhe Chen, Guo Chen, Wenhai Wang, and Tong Lu,
\newblock ``Avsegformer: Audio-visual segmentation with transformer,'' 2023.

\bibitem{wang2023loconet}
Xizi Wang, Feng Cheng, Gedas Bertasius, and David Crandall,
\newblock ``Loconet: Long-short context network for active speaker detection,''
\newblock {\em arXiv preprint arXiv:2301.08237}, 2023.

\bibitem{LiuSMDAFAS}
Jiawei Liu, Wayne Lam, Zhigang Zhu, and Hao Tang,
\newblock ``Smdaf: A scalable sidewalk material data acquisition framework with bidirectional cross-modal knowledge distillation,''
\newblock .

\bibitem{gan2019self}
Chuang Gan, Hang Zhao, Peihao Chen, David Cox, and Antonio Torralba,
\newblock ``Self-supervised moving vehicle tracking with stereo sound,''
\newblock in {\em Proceedings of the IEEE/CVF international conference on computer vision}, 2019, pp. 7053--7062.

\bibitem{riverahurtado2021mmdistillnet}
Francisco Rivera~Valverde, Juana Valeria~Hurtado, and Abhinav Valada,
\newblock ``There is more than meets the eye: Self-supervised multi-object detection and tracking with sound by distilling multimodal knowledge,''
\newblock 2021.

\bibitem{vasudevan2020semantic}
Arun~Balajee Vasudevan, Dengxin Dai, and Luc Van~Gool,
\newblock ``Semantic object prediction and spatial sound super-resolution with binaural sounds,''
\newblock in {\em European conference on computer vision}. Springer, 2020, pp. 638--655.

\bibitem{Mao2023MultimodalVA}
Yuxin Mao, Jing Zhang, Mochu Xiang, Yiran Zhong, and Yuchao Dai,
\newblock ``Multimodal variational auto-encoder based audio-visual segmentation,''
\newblock {\em 2023 IEEE/CVF International Conference on Computer Vision (ICCV)}, pp. 954--965, 2023.

\bibitem{Gan2019SelfSupervisedMV}
Chuang Gan, Hang Zhao, Peihao Chen, David~D. Cox, and Antonio Torralba,
\newblock ``Self-supervised moving vehicle tracking with stereo sound,''
\newblock {\em 2019 IEEE/CVF International Conference on Computer Vision (ICCV)}, pp. 7052--7061, 2019.

\bibitem{Liu2019StructuredKD}
Yifan Liu, Changyong Shun, Jingdong Wang, and Chunhua Shen,
\newblock ``Structured knowledge distillation for dense prediction,''
\newblock {\em IEEE Transactions on Pattern Analysis and Machine Intelligence}, vol. 45, pp. 7035--7049, 2019.

\bibitem{Wang2020PayAT}
Kafeng Wang, Xitong Gao, Yiren Zhao, Xingjian Li, Dejing Dou, and Chengzhong Xu,
\newblock ``Pay attention to features, transfer learn faster cnns,''
\newblock in {\em International Conference on Learning Representations}, 2020.

\bibitem{TowardsRobustAudioBasedVehicleDetectio}
Jung~Uk Kim and Seong Tae~Kim,
\newblock ``Towards robust audio-based vehicle detection via importance-aware audio-visual learning,''
\newblock in {\em ICASSP 2023 - 2023 IEEE International Conference on Acoustics, Speech and Signal Processing (ICASSP)}, 2023, pp. 1--5.

\end{thebibliography}

\end{document}